\documentclass[11pt,a4paper]{article}
\usepackage[hyperref]{acl2020}
\usepackage{times}
\usepackage{latexsym}
\usepackage{amsfonts}
\usepackage{graphicx}
\usepackage{multirow}
\usepackage{color,xcolor}
\usepackage{enumitem}
\usepackage{placeins}

\usepackage{float}
\usepackage{multicol}
\usepackage{CJKutf8}

\usepackage{microtype}

\aclfinalcopy 

\title{Integrating Semantic and Structural Information with Graph Convolutional Network for Controversy Detection}

\author{Lei Zhong$^{1,2}$,  Juan Cao$^{1,2}$\thanks{$^*Corresponding \ author.$}, Qiang Sheng$^{1,2}$, Junbo Guo$^{1}$, Ziang Wang$^{1,2}$  
\\
 \textsuperscript{1}Key Laboratory of Intelligent Information Processing of Chinese Academy of Sciences (CAS) \\
 \& Center for Advanced Computing Research, \\ Institute of Computing Technology, CAS, Beijing, China \\
  \textsuperscript{2}University of Chinese Academy of Sciences, Beijing, China\\
  \texttt{\{zhonglei18s, caojuan, shengqiang18z, guojunbo\}@ict.ac.cn} \\\texttt{gnaizgnaw@gmail.com} 
}

\date{}

\begin{document}

\maketitle


\begin{abstract}
Identifying controversial posts on social media is a fundamental task for mining public sentiment, assessing the influence of events, and alleviating the polarized views. However, existing methods fail to 1) effectively incorporate the semantic information from content-related posts; 2) preserve the structural information for reply relationship modeling; 3) properly handle posts from topics dissimilar to those in the training set. To overcome the first two limitations, we propose \underline{T}opic-\underline{P}ost-\underline{C}omment \underline{G}raph \underline{C}onvolutional \underline{N}etwork (TPC-GCN), which integrates the information from the graph structure and content of topics, posts, and comments for post-level controversy detection. As to the third limitation, we extend our model to Disentangled TPC-GCN (DTPC-GCN), to disentangle topic-related and topic-unrelated features and then fuse dynamically.
Extensive experiments on two real-world datasets demonstrate that our models outperform existing methods. Analysis of the results and cases proves that our models can integrate both semantic and structural information with significant generalizability.
\end{abstract}

\section{Introduction}
\label{sec:intro}

\begin{figure}[ht]
	\centering
	\includegraphics[width=0.46\textwidth]{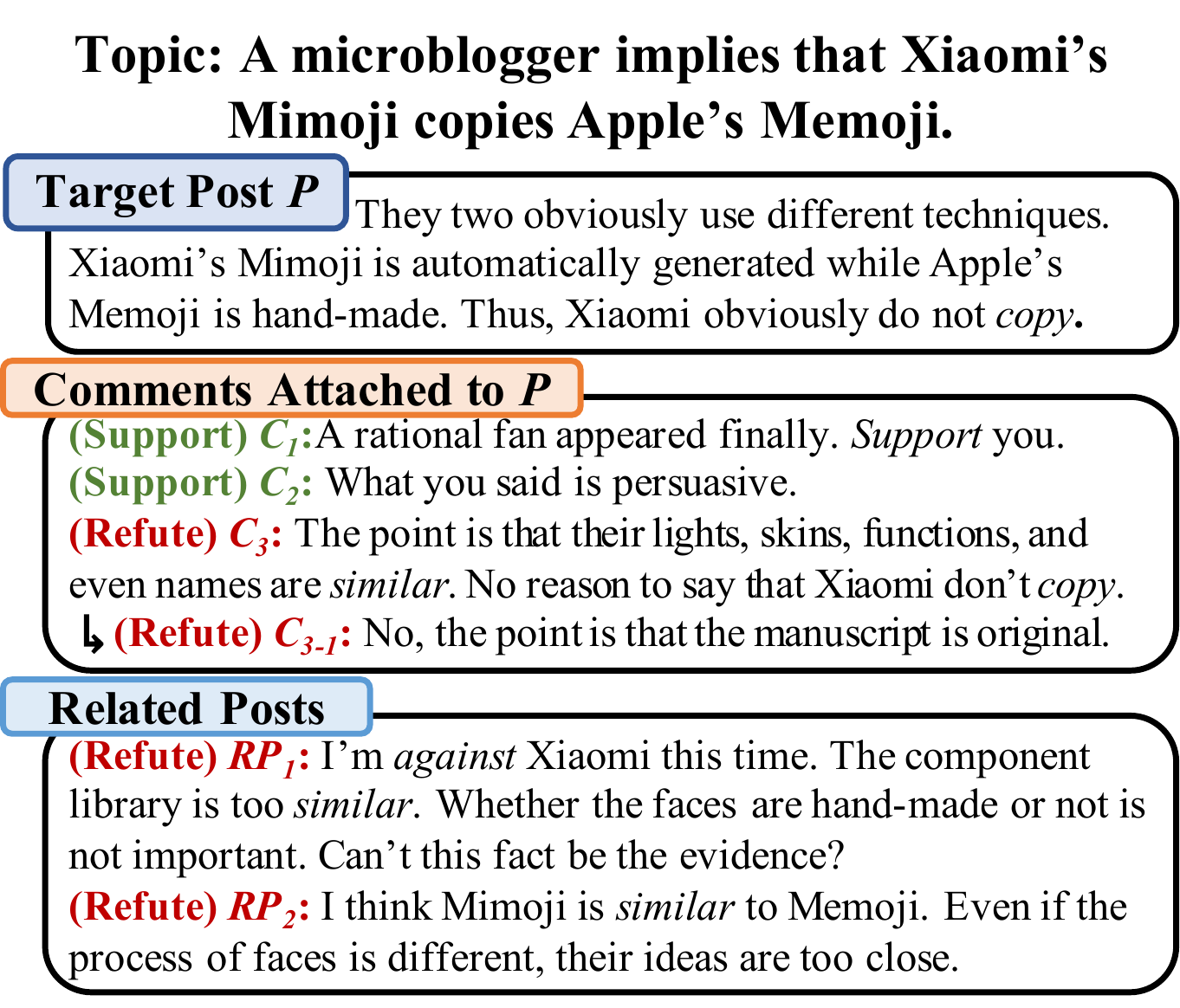}
	\caption{A controversial post $P$ about whether Xiaomi's Mimoji copies Apple's Memoji. These {\color[rgb]{0.275,0.510,0.231} \textbf{Supports}} and {\color[rgb]{0.788,0.008,0.075} \textbf{Refutations}} are to either their respective parent comments or $P$.}
	\label{fig:first_eg}
\end{figure}	

Social media such as Reddit\footnote{\url{https://www.reddit.com/}} and Chinese Weibo\footnote{\url{https://weibo.com/}} has been the major channel through which people can easily propagate their views. In the open and free circumstance, the views expressed by the posts often spark fierce discussion and raise controversy among the engaging users. These controversial posts provide a lens of public sentiment, which bring about several tasks such as news topic selection, influence assessment \cite{hessel2019something}, and alleviation of polarized views \cite{garimella2017reducing}. As a basis of all mentioned tasks, automatically identifying the controversial posts has attracted wide attention \cite{addawood2017telling,coletto2017motif,rethmeier2018learning,hessel2019something}.

This work focuses on post-level controversy detection on social media, i.e., to classify if a post is controversial or non-controversial. According to \cite{coletto2017motif}, a controversial post has debatable content and expresses an idea or an opinion which generates an argument in the responses, representing opposing opinions in favor or in disagreement with the post. In practice, the responses of a target post (the post to be judged) generally come from two sources, i.e., the comments attached to the post and other content-related posts. Figure \ref{fig:first_eg} shows an example where the target post $P$ expresses that Xiaomi's Mimoji do not copy Apple's Memoji. We can see that: 1) The comments show more supports and fewer refutes to $P$, which raises a small controversy. However, the related posts show extra refutations and enhance the controversy of $P$. 2) $C_{3-1}$ expresses refutation literally, but it actually supports $P$ because in the comment tree, it refutes $C_3$, a refuting comment to $P$. 3) There exist two kinds of semantic clues for detection, topic-related and topic-unrelated clues. For example, \textit{support} and \textit{against} is unrelated to this topic, while \textit{copy} and \textit{similar} are topic-related. Topic-related clues can help identify posts in a similar topic, but how effective they are for those in dissimilar topics depends on the specific situation. Therefore, to comprehensively evaluate the controversy of a post, the information from both the comments and related posts should be integrated properly on semantic and structure level.

Existing methods detecting controversy on social media have exploited the semantic feature of the target post and its comments as well as structural feature. However, three drawbacks limit their performance: 1) These methods ignore the role of the related posts in the same topic in providing extra supports or refutations on the target post. Only exploiting the information from comments is insufficient. 2) These methods use statistical structure-based features which cannot model the reply-structure relationships (like $P$-$C_1$ and $C_3$-$C_{3-1}$ in Figure \ref{fig:first_eg}). The stances of some comments may be misunderstood by the model (like $C_{3-1}$). 3) These methods tend to capture topic-related features that are not shared among different topics with directly using information of content \cite{wang2018eann}. The topic-related features can be helpful when the testing post is from a topic similar to those in the training set but would hurt the detection otherwise.
	
\begin{figure*}[ht]
	\centering
	\includegraphics[width=0.7\textwidth]{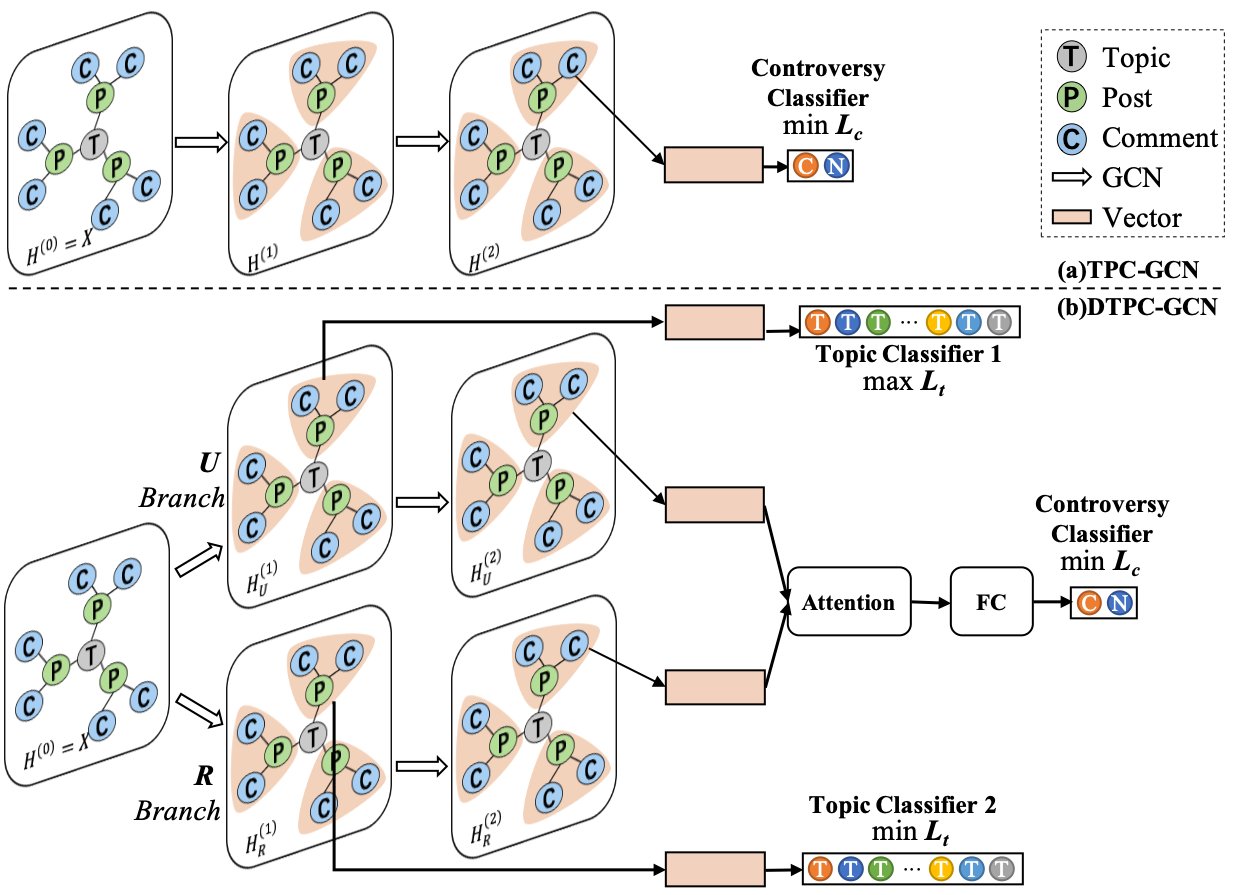}
	\caption{Architecture of (a) Topic-Post-Comment Graph Convolutional Network (TPC-GCN).  (b) Disentangled TPC-GCN (DTPC-GCN). The upper post in the TPC graph is taken as an example to illustrate the methods. $H_B^{(l)}$ is the representation matrix, containing all node vectors in the $l$-th layer of Branch $B$. $X$ is the initial representation. $L_c$ and $L_t$ refer to controversy classification loss and topic classification loss respectively. FC means fully connected layer.}
	\label{fig:model}
\end{figure*}

Recently, graph convolutional networks have achieved great success in many areas \cite{marcheggiani2018exploiting, ying2018graph, yao2019graph, li2019encoding} due to its ability to encode both local graph structure and features of node \cite{kipf2016semi}. To overcome the first two drawbacks of existing works, we propose a \underline{T}opic-\underline{P}ost-\underline{C}omment \underline{G}raph \underline{C}onvolutional \underline{N}etwork (TPC-GCN) (see Figure \ref{fig:model}a) that integrates the information from the graph structure and content of topics, posts, and comments for post-level controversy detection. First, we create a TPC graph to describe the relationship among topics, posts, and comments. 
To preserve the reply-structure information, we connect each comment node with the post/comment node it replies to. To include the information from related posts, we connect each post node with its topic node. 
Then, a GCN model is applied to learn node representation with content and reply-structure information fused.
Finally, the updated vectors of a post and its comments are fused to predict the controversy.

TPC-GCN is mainly for detection in intra-topic mode, i.e., topics of testing posts appear in the training set, for it cannot overcome the third drawback. We thus extend a two-branch version of TPC-GCN named \underline{D}isentangled TPC-GCN (DTPC-GCN) (see Figure \ref{fig:model}b) for inter-topic mode (no testing posts are from the topics in the training set). We use a TPC-GCN in each branch, but add an auxiliary task, topic classification. The goals of the two branches for the auxiliary task are opposite to disentangle the topic-related and topic-unrelated features. The disentangled features can be dynamically fused according to the content of test samples with attention mechanism for final decision. Extensive experiments demonstrate that our models outperform existing methods and can exploit features dynamically and effectively. The main contributions of this paper are as follows:

\begin{enumerate}
	\item We propose two novel GCN-based models, TPC-GCN and DTPC-GCN,  for post-level controversy detection. The models can integrate the information from the structure and content of topics, posts, and comments, especially the information from the related posts and reply tree. Specially, DTPC-GCN can further disentangle the topic-related features and topic-unrelated features for inter-topic detection.
    \item We build a Chinese dataset for controversy detection, consisting of 5,676 posts collected from Chinese Weibo, each of which are manually labeled as controversial or non-controversial. To the best of our knowledge, this is the first released Chinese dataset for controversy detection.
	\item Experiments on two real-world datasets demonstrate that the proposed models can effectively identify the controversial posts and outperform existing methods in terms of performance and generalization.
\end{enumerate}

\section{Related Work}

Controversy detection on the Internet have been studied on both web pages and social media. Existing works detecting controversy on web pages mostly aims at identifying controversial articles in Wikipedia. Early methods are mainly based on statistical features, such as revision times \cite{kittur2007he}, edit history \cite{vuong2008ranking,yasseri2012dynamics,rad2012identifying} and dispute tag \cite{dori2015automated}. Others incorporate the collaboration-network-based features, sentiment-based features \cite{vuong2008ranking, wang2016piece}, and semantic features \cite{linmans2018improved}. As to the common web pages, existing works exploit the controversy on Wikipedia \cite{awadallah2012harmony,dori2013detecting,dori2015automated,jang2016probabilistic} and user comments \cite{choi2010identifying,tsytsarau2011scalable} for detection.

Unlike the web pages, social media contains more diverse topics and more fierce discussion among users, which makes controversy detection on social media more challenging. Early studies assume that a topic has its intrinsic controversy, and focus on topic-level controversy detection. \citet{popescu2010detecting} detect controversial snapshots (consisting of many tweets referring to a topic) based on Twitter-based and external-knowledge features.  \citet{garimella2018quantifying} build graphs based on a Twitter topic, such as retweeting graph and following graph, and then apply graph partitioning to measure the extent of controversy. However, topic-level detection is rough, because there exists non-controversial posts in a controversial topic and vice versa. Recent works focus on post-level controversy detection by leveraging language features, such as emotional and topic-related phrases \cite{rethmeier2018learning}, emphatic features, Twitter-specific features \cite{addawood2017telling}. Other graph-based methods exploit the features from the following graph and comment tree \cite{coletto2017motif, hessel2019something}. The limitations of current post-level works are that they do not effectively integrate the information from content and reply-structure, and ignore the role of posts in the same topic. Moreover, the difference between intra-topic and inter-topic mode is not realized. Only \citet{hessel2019something} deal with topic transfer, but they train on each topic and test on others to explore the transferability, which is not suitable in practice.

\section{Methodology}

In this section, we introduce the Topic-Post-Comment Graph Convolutional Network (TPC-GCN) and its extension Disentangled TPC-GCN (DTPC-GCN), as shown in Figure \ref{fig:model}. We first introduce the TPC graph construction and then detail the two models.

\subsection{TPC Graph Construction}
\label{ref:graph_construction}
To model the paths of message passing among topics, posts, and comments, we first construct a topic-post-comment graph $G=(V,E)$ for target posts, where $V$ and $E$ denote the set of nodes and edges respectively. 
First, to preserve the post-comment and inter-comment relationship, we incorporate the comment tree, each comment node of which is connected with the post/comment node it replies to. Then, to facilitate the posts capturing information from related posts in the same topic that proved helpful in Section \ref{sec:intro}, we connect each post with its topic. The topic node can be regarded as a \textit{hub} node to integrate and interchange the information. Another way is to connect post nodes in a topic pairwise, but the complexity will be high. Note that the concept \textit{topic} here is not necessarily provided by the platform, such as the subreddit on Reddit and the hashtag (\#) on Weibo. When topics are not provided, algorithms for text-based clustering can be used to construct a topic with related posts \cite{nematzadeh2019empirical}.

In $G$, each node may represent a topic, a post, or a comment and each edge may represent topic-post, post-comment, or comment-comment connection. We initially represent each node $v$ with an embedding vector $x$ of their text by using the pre-trained language model.

\subsection{TPC-GCN}
In this subsection, we detail the TPC-GCN, by first introducing the generic GCN and then our TPC-GCN model.

The GCN has been proved an efficient neural network that operates on a graph to encode both local graph structure and features of node \cite{kipf2016semi}. The characteristic of GCN is consistent to our goal that integrates the semantic and structural information. In a GCN, each node is updated according to the aggregated information of its neighbor nodes and itself, so the learned representation can include information from both content and structure. For a node $v_i \in V$, the update rule in the message passing process is as follows:
\begin{equation}
	\label{eq:nodegcn}
    h_i^{(l+1)} = \sigma \left( \sum_{j\in N_i} g \left(h_i^{(l)}, h_j^{(l)} \right) + b^{(l)} \right)
\end{equation}
where $h_i^{(l)}$ is the hidden state of node $v_i$ in the $l$-th layer of a GCN
 and $N_i$ is the neighbor set of node $v_i$ with itself included. Incoming messages from $N_i$ are transformed by the function $g$ and then pass through the activation function $\sigma$ (such as \texttt{ReLU}) to output new representation for each node. $b^{(l)}$ is the bias term. Following \citet{kipf2016semi}, we use a linear transform function $g(h_i^{(l)}, h_j^{(l)}) = W^{(l)}h_j$, where $W^{(l)}$ is a learnable weight matrix. Based on node-wise Equation \ref{eq:nodegcn}, layer-wise propagation rule can be written as the following form:
\begin{equation}
	\label{eq:layergcn}
    H^{(l+1)} = \sigma \left( \hat{A}H^{(l)}W^{(l)} + B^{(l)}\right)
\end{equation}
where $H^{(l)}$ contains all node vectors in the $l$-th layer and $\hat{A}$ is the normalized adjacency matrix with inserted self-loops. $W^{(l)}$ is the weight matrix and $B^{(l)}$ is the broadcast bias term.

In TPC-GCN (see Figure \ref{fig:model}a), we input the matrix consisting of $N$ $d$-dimensional embedding vectors $H^{(0)} = X \in \mathbb{R}^{N \times d}$ to a two-layer GCN to obtain the representation after message passing $H^{(2)}$. Next, the vector of each post node $i$ and its attached comment nodes are averaged to be the fusion vector $f_i$ of the post. Finally, we apply a \texttt{softmax} function to the fusion vectors for the controversy probability of each post. The cross entropy is the loss function:
\begin{equation}
	\label{eq:closs}
	L_c = -\frac{1}{N}\sum_i \left( \left( 1-y^c_i \right)\log\left(1-p^c_i\right) + y^c_i\log \left(p^c_i\right)\right) 
\end{equation}
where $y^c_{i}$ is a label with 1 representing \textit{controversial} and 0 representing the \textit{non-controversial}, $p^c_{i}$ is the predicted probability that the $i$-th post is controversial, and $N$ is the size of training set. The limit of TPC-GCN is that the representation tends to be topic-related as Section \ref{sec:intro} said. The limited generalizability of TPC-GCN makes it more suitable for intra-topic detection, instead of inter-topic detection.

\subsection{Disentangled TPC-GCN}
Intuitively, topic-unrelated features are more effective when testing on the posts from unknown topics (inter-topic detection). However, topic-related features can help when unknown topics are similar to the topics in the training set. Therefore, both of topic-related and topic-unrelated features are useful, but their weights vary from sample to sample. This indicates that the two kinds of features should be disentangled and then dynamically fused. Based on the above analysis, we propose the extension of TPC-GCN, Disentangled TPC-GCN (see Figure \ref{fig:model}b), for inter-topic detection. DTPC-GCN consists of two parts: the two-branch multi-task architecture for disentanglement, and attention mechanism for dynamic fusion.

\noindent\textbf{Two-branch Multi-task Architecture } 
To obtain the topic-related and topic-unrelated features at the same time, we use two branches of TPC-GCN with multi-task architecture, denoted as $R$ for topic-related branch and $U$ for topic-unrelated one. In both $R$ and $U$, an auxiliary task, topic classification, is introduced to guide the learning of representation oriented by the topic.

For each branch, we first train the first layer of GCN with the topic classification task. The input of the topic classifier is fusion vectors from $H^{(1)}$ which are obtained with the same process of $f_i$ in TPC-GCN. The cross entropy is used as the loss function:
\begin{equation}
	\label{eq:rloss}
	L_t = -\frac{1}{N} \sum_{k}\sum_{i} y^t_{ik}\log(p^t_{ik})
\end{equation}
where $y^t_{ik}$ is a label with 1 representing the ground-truth topic and 0 representing the incorrect topic class, $p^t_{ik}$ is the predicted probability of the $i$-th post belonging to the $k$-th topic, and $N$ is the size of training set. The difference between $R$ and $U$ is that we minimize $L_t$ in Branch $R$ to obtain topic-distinctive features, but maximize $L_t$ in Branch $U$ to obtain topic-confusing features.

Then we include the second layer of GCN and train on two tasks, i.e., topic and controversy classification, for each branch individually. Branch $U$ and $R$ are expected to evaluate controversy effectively with different features in terms of the relationship with the topics.

\noindent\textbf{Attention Mechanism }
After the individual training, Branch $U$ and $R$ are expected to capture the topic-related and topic-unrelated features respectively. We further fuse the features from the two branches dynamically. Specifically, we freeze the parameters of $U$ and $R$, and further train the dynamic fusion component. For the weighted combination of fusion vectors $f_{U}$ and $f_{R}$ from the two branches, we use the attention mechanism as follows:
\begin{equation}
	\mathcal{F}(f_{b}) = v^T \tanh(W_\mathcal{F} f_{b} + b_\mathcal{F}), b \in\{U,R\}
\end{equation}
\begin{equation}
	\alpha_{b} = \frac{\exp(\mathcal{F}(f_{b}))}{\sum_{b\in \{U,R\}}\exp(\mathcal{F}(f_{b}))}
\end{equation}
\begin{equation}
	u = \sum_{b\in\{U,R\}} \alpha_{b}f_b
\end{equation}
where $W_{\mathcal{F}}$ is the weight matrix and $b_{\mathcal{F}}$ is the bias term. $v^T$ is a transposed weight vector and $\mathcal{F}(\cdot)$ outputs the score of the input vector. The scores of features from Branch $U$ and $R$ are normalized via a \texttt{softmax} function as the branch weight. The weighted sum of the two fusion vectors $u$ is finally used for controversy classification. The loss function is the same as Equation \ref{eq:closs}.

\section{Experiment}

In this section, we conduct experiments to compare our proposed models and other baseline models. Specifically, we mainly answer the following evaluation questions:

\noindent\textbf{EQ1:} Are TPC-GCN and DTPC-GCN able to improve the performance of controversy detection?

\noindent\textbf{EQ2:} How effective are different information in TPC-GCN, including the content of topics, posts, and comments as well as the topic-post-comment structure?

\noindent\textbf{EQ3:} Can DTPC-GCN learn disentangled features and dynamically fuse them for controversy detection?

\subsection{Dataset}

\begin{table}
\centering
\small
\setlength{\tabcolsep}{3pt}
\begin{tabular}{lll}
\hline \textbf{Number} & \textbf{Weibo} & \textbf{Reddit} \\ \hline
Topics(Hashtags/Subreddits) & 49 & 6\\
Controversial Posts  & 1,992 & 7,515 \\
Non-controversial Posts & 3,684 & 7,518  \\
All Posts & 5,676 & 15,033 \\
Comments of Controversial Posts & 35,632 & 578,879 \\
Comments of Non-Controversial Posts & 34,565 & 1,461,697\\
All Comments & 70,197 & 2,040,576 \\
\hline
\end{tabular}
\caption{\label{table:dataset} Statistics of two datasets. }
\end{table}

We perform our experiments on two real-world datasets in different languages. Table \ref{table:dataset} shows the statistics of the two datasets. The details are as follows:

\noindent\textbf{Reddit Dataset } The Reddit dataset released by \citet{hessel2019something} and Jason Baumgartner of \url{pushshift.io} is the only accessible English dataset for controversy detection of social media posts. This dataset contains six subreddits (which can be regarded as over-arching topics): $\mathrm{AskMen}$, $\mathrm{AskWomen}$, $\mathrm{Fitness}$, $\mathrm{LifeProTips}$, $\mathrm{personalfinance}$, and $\mathrm{relationships}$. Each post belongs to a subreddit and the number of attached comments is ensured to be over 30. The tree structure of the comments is also maintained. We use the comment data in the first hour after a post is published.

\noindent\textbf{Weibo Dataset } We built a Chinese dataset for controversy detection on Weibo \footnote{\url{http://mcg.ict.ac.cn/controversy-detection-dataset.html}} in this work. We first manually selected 49 widely discussed, multi-domain topics from July 2017 to August 2019 (see Appendix \ref{append:topics}). Then, we crawled the posts on those topics and preserved those with at least two comments. Here we rebuilt the comment tree according to the comment time and usernames due to the lack of officially-provided structure. Finally, annotators were asked to read and then annotate the post based on both of the post content and the user stances in the comments/replies. Each post was labeled by two annotators(Cohen's Kappa coefficient = 0.71). When the disagreement occurred between the annotators, the authors discussed and determined the labels. In total, this dataset contains 1,992 controversial posts and 3,684 non-controversial posts, which is in line with the distribution imbalance in the real-world scenario. As far as we know, this is the first released dataset for controversy detection on Chinese social media. We use at most 15 comments of each post due to the computation limit.

In the intra-topic experiment: For the Weibo dataset, we randomly divided with a ratio of 4:1:1 in each topic and merged them respectively across all topics. For the Reddit dataset, we apply the data partition provided by the authors. The ratio is 3:1:1.

In the inter-topic experiments: For the Weibo and Reddit dataset, we still divided with a ratio of 4:1:1, but on the topic level.

\subsection{Implementation Details}

In the (D)TPC-GCN model, each node is initialized with its textual content using the pre-trained BERT\footnote{\url{https://github.com/google-research/bert}} (BERT-Base Chinese for Weibo and BERT-Base Uncased for Reddit) and the padding size for each is 45. We only fine-tune the last layer, namely layer 11 of BERT for simplicity and then apply a dense layer with a \texttt{ReLU} activation function to reduce the dimensionality of representation from 768 to 300. In TPC-GCN, the sizes of hidden states of the two GCN layers are 100 and 2, respectively, with \texttt{ReLU} for the first GCN layer. To avoid overfitting, a dropout layer is added between the two layers with a rate of 0.35. We apply a \texttt{softmax} function to the fusion vector for obtaining the controversy probability. In DTPC-GCN, the size of hidden states of the first and second GCN layers in each branch are 32 and 16. The dropout rate between two GCN layers in each branch is set to 0.4. The batch size in our (D)TPC-GCN model is 1 (1 TPC graph), and 128 (posts and attached replies) in our PC-GCN model and baselines. The optimizer is BertAdam\footnote{\url{https://pypi.org/project/pytorch-pretrained-bert/}} in all BERT-based models and Adam \cite{kingma2014adam} in the other semantic models. The learning rate is 1e-4 and the total epoch is 100. We report the best model according to the performance on the validation set. In those semantic models that are not based on BERT, we use two publicly-available big-scale word embedding files to obtain the model input, \textit{sgns.weibo.bigram-char}\footnote{\url{https://github.com/Embedding/Chinese-Word-Vectors}} for Weibo and \textit{glove.42B.300d}\footnote{\url{https://nlp.stanford.edu/projects/glove/}} for Reddit.

\begin{table*}[h]
\centering
\small
\setlength{\tabcolsep}{3pt}
\begin{tabular}{ll|cccc|cccc}
\hline 
\multicolumn{2}{c|}{\multirow{2}{*}{\textbf{Method}}}  & \multicolumn{4}{c|}{\textbf{Weibo Dataset}} & \multicolumn{4}{c}{\textbf{Reddit Dataset}} \\
\cline{3-10} & & \textbf{Avg. P} & \textbf{Avg. R} & \textbf{Avg. F1} & \textbf{Acc.} & \textbf{Avg. P} & \textbf{Avg. R} & \textbf{Avg. F1} & \textbf{Acc.} \\ \hline
\multirow{4}{*}{\textbf{Content-based}} 
 & TextCNN & 72.80 & 68.49 & 69.08 & 72.83 & 56.58 & 56.33 & 55.92 & 56.33 \\
 & BiLSTM-Att & 69.97 & 70.31 & 70.10 & 71.28 & 62.74 & 60.66 & 58.98 & 60.66\\
 & BiGRU-Att & 71.35 & 72.21 & 71.50 & 72.21 & 59.95 & 59.86 & 59.77 & 59.86\\
 & BERT &	72.17 &	72.72 &	72.37 &	73.35 & 60.80 & 60.80 & 60.80 & 60.80\\
\hline
\textbf{Structure-based} & SFC & 68.15 & 66.27 & 66.72 & 70.10 & 59.47 & 59.47 & 59.47 & 59.47 \\
\hline
\multirow{2}{*}{\textbf{Fusion}}
& \cite{hessel2019something} & 72.52 & 70.82 & 71.34 & 73.82 & 63.03 & 63.03 & 63.03 & 63.03 \\
& TPC-GCN & \textbf{74.65} & \textbf{75.33} & \textbf{74.88} & \textbf{75.72}  & \textbf{67.00} & \textbf{66.97} & \textbf{66.95} & \textbf{66.97}\\
\hline
\end{tabular}
\caption{\label{table:intratopic} Performance(\%) comparison of the intra-topic experiments.}
\end{table*}

\begin{table*}[h]
\centering
\small
\setlength{\tabcolsep}{3pt}
\begin{tabular}{ll|cccc|cccc}
\hline 
\multicolumn{2}{c|}{\multirow{2}{*}{\textbf{Method}}}  & \multicolumn{4}{c|}{\textbf{Weibo Dataset}} & \multicolumn{4}{c}{\textbf{Reddit Dataset}} \\
\cline{3-10} & & \textbf{Avg. P} & \textbf{Avg. R} & \textbf{Avg. F1} & \textbf{Acc.} & \textbf{Avg. P} & \textbf{Avg. R} & \textbf{Avg. F1} & \textbf{Acc.} \\ \hline
\multirow{4}{*}{\textbf{Content-based}} 
 & TextCNN & 71.55 & 72.63 & 69.63 & 69.76 & 54.20 & 54.18 & 54.12 & 54.18 \\
 & BiLSTM-Att & 67.09 & 68.09 & 67.10  & 68.00 & 60.96 & 59.76 & 58.63  & 59.76 \\
 & BiGRU-Att & 68.04 & 67.08 & 67.35 & 70.18 & 58.49 & 58.17 & 57.76 & 58.17\\
 & BERT & 68.77 & 68.16 &	68.42 & 72.22 & 60.41 & 59.96 & 59.53 & 59.96 \\
\hline
\textbf{Structure-based} & SFC & 63.06 & 63.69 & 63.04 & 64.03 & 58.87 & 58.86 & 58.86 & 58.86 \\
\hline
\multirow{3}{*}{\textbf{Fusion}}
& \cite{hessel2019something} & 69.25  & 67.15 & 67.63 & 70.84  & 60.77 & 60.76 & 60.74 & 60.76 \\
& TPC-GCN & 73.84 & 72.00 & 71.53 & 72.11  & 63.39 & 63.24 & 63.14 & 63.24 \\
& DTPC-GCN &\textbf{75.57} & \textbf{75.31} & \textbf{75.27} & \textbf{75.35}& \textbf{68.76} & \textbf{67.63} & \textbf{67.14} & \textbf{67.63}\\
\hline
\end{tabular}
\caption{\label{table:intertopic} Performance(\%) comparison of the inter-topic experiments.}
\end{table*}

\subsection{Baselines}

To validate the effectiveness of our methods, we implemented several representative methods including content-based, structure-based and fusion methods as baselines.

\noindent\textbf{Content-based Methods }

We implement mainstream text classification models including \textbf{TextCNN} \cite{kim2014convolutional}, \textbf{BiLSTM-Att} (bi-directional LSTM with attention) BiLSTM \cite{graves2005framewise,bahdanau2014neural}, \textbf{BiGRU-Att} (bi-directional GRU with attention) \cite{cho2014properties},\textbf{BERT} \cite{devlin2018bert} (only fine-tune the last layer for simplicity). For a fair comparison, we concatenate the post and its attached comments together as the input, instead of feeding the post only.

\noindent\textbf{Structure-based Methods}

Considering that structure-based features of the post and its comment tree are rare and non-systematic in previous works, we integrate the plausible features in \cite{coletto2017motif} and \cite{hessel2019something}. As the latter paper does, we feed them into a series of classifiers and choose a best model for classification. We name the method \textbf{SFC}. For a post-comment graph, the feature set contains the average depth (average length of root-to-leaf paths), the maximum relative degree (the largest node degree divided by the degree of the root), C-RATE features (the logged reply time between the post and comments, or over pairs of comments), and C-TREE features (statistics in a comment tree, such as maximum depth/total comment ratio).

\noindent\textbf{Fusion Method} 

The compared fusion method from \cite{hessel2019something} aims to identify the controversial posts with semantic and structure information. They extract text features of topics, posts, and comments by BERT and structural feature including the C-RATE and C-TREE features mentioned above. In addition, publish time features are also exploited.

\subsection{Performance Comparison}

 To answer \textbf{EQ1}, we compare the performance of proposed (D)TPC-GCN with mentioned baselines on the two datasets. The evaluation metrics include the macro average precision (Avg. P), macro average recall (Avg. R), macro average F1 score (Avg. F1), and accuracy (Acc.). Table \ref{table:intratopic} and \ref{table:intertopic} show the performance of all compared methods for intra-topic detection and inter-topic detection respectively.
 
 In the intra-topic experiments, we can see that 1) TPC-GCN outperforms all compared methods on the two datasets. This indicates that our model can effectively detect controversy with a significant generalizability on different datasets.
  2) The structure-based model, SFC, reports the low scores on the two datasets, indicating that the statistical structural information is insufficient to timely identify the controversy.
  3) The fusion models outperform or are comparable to the other baselines, which proves that information fusion of content and structure is necessary to improve the performance.
  
 In the inter-topic experiments, we can see that 1) DTPC-GCN outperforms all baselines by 6.4\% of F1 score at least, which validates that DTPC-GCN can detect controversy on unseen or dissimilar topics. 2) DTPC-GCN outperforms TPC-GCN by 3.74\% on Weibo and 4.00\% on Reddit. This indicates that feature disentanglement and dynamic fusion can significantly improve the performance of inter-topic controversy detection.
 


\begin{table*}
\centering
\small
\setlength{\tabcolsep}{3pt}
\begin{tabular}{l|cccc|cccc}
\hline 
\multirow{2}{*}{\textbf{Method}}  & \multicolumn{4}{c|}{\textbf{Weibo Dataset}} & \multicolumn{4}{c}{\textbf{Reddit Dataset}} \\
\cline{2-9} & \textbf{Avg. P} & \textbf{Avg. R} & \textbf{Avg. F1} & \textbf{Acc.} & \textbf{Avg. P} & \textbf{Avg. R} & \textbf{Avg. F1} & \textbf{Acc.} \\ \hline
TPC-GCN &  \textbf{74.65} & \textbf{75.33} & \textbf{74.88} & \textbf{75.72} & \textbf{67.00} & \textbf{66.97} & \textbf{66.95} & \textbf{66.97} \\
\hline
PC-GCN & 73.49 & 74.16 & 73.72 & 74.59 & 66.48 & 65.60 & 65.14 & 65.60 \\
TP-GCN & 58.72 & 59.16 & 58.20 & 58.68 & 52.97 & 52.83 & 52.28 & 52.83 \\
\hline
(RT)PC-GCN & 71.78 & 71.07 & 71.35 & 73.14 & 65.86 & 65.80 & 65.77 & 65.80 \\
T(RP)C-GCN & 72.30 & 72.65 & 72.45 & 73.55 & 65.25 & 64.73 & 64.43 & 64.73 \\
TP(RC)-GCN & 59.66 & 59.80 & 59.71 & 61.36 & 62.98 & 62.80 & 62.67 & 62.80 \\
 
\hline
\end{tabular}
\caption{\label{ablation study:intra} Ablation study of TPC-GCN in the intra-topic experiments (\%).}
\end{table*}

\begin{table*}
\centering
\small
\setlength{\tabcolsep}{3pt}
\begin{tabular}{l|cccc|cccc}
\hline 
\multirow{2}{*}{\textbf{Method}}  & \multicolumn{4}{c|}{\textbf{Weibo Dataset}} & \multicolumn{4}{c}{\textbf{Reddit Dataset}} \\
\cline{2-9} & \textbf{Avg. P} & \textbf{Avg. R} & \textbf{Avg. F1} & \textbf{Acc.} & \textbf{Avg. P} & \textbf{Avg. R} & \textbf{Avg. F1} & \textbf{Acc.} \\ \hline
DTPC-GCN & \textbf{75.57} & \textbf{75.31} & \textbf{75.27} & \textbf{75.35} & \textbf{68.76} & \textbf{67.63} & \textbf{67.14} & \textbf{67.63}\\
U branch only & 74.06 & 74.06 & 74.05 & 74.05 &  63.95 & 63.94 & 63.94 & 63.94\\
R branch only & 74.16 & 73.33 & 73.15 & 73.41 & 63.41 & 63.15 & 62.97 & 63.15\\
\hline
\end{tabular}
\caption{\label{ablation study:inter} Ablation study of DTPC-GCN in the  inter-topic experiments (\%).}
\end{table*}

\subsection{Ablation Study}

To answer \textbf{EQ2} and part of \textbf{EQ3}, we also evaluate several internal models, i.e., the simplified variations of (D)TPC-GCN by removing some components or masking some representations. By the ablation study, we aim to investigate the impact of content and structural information in TPC-GCN and topic-related and topic-unrelated information in DTPC-GCN.

\noindent\textbf{Ablation Study of TPC-GCN}

We delete certain type of nodes (and the edges connect to them) to investigate their overall impact and mask the content by randomizing the initial representation to investigate the impact of content. 
Specifically, we investigate on the following simplified models of TPC-GCN:

\textbf{PC-GCN} / \textbf{TP-GCN}: discard the topic / comment nodes.

\textbf{(RT)PC-GCN} / \textbf{T(RP)C-GCN} / \textbf{TP(RC)-GCN}: randomly initialize the representation of topic / post / comment nodes.

From Table \ref{ablation study:intra}, we have the following observations: 
1) TPC-GCN outperforms all simplified models, indicating that the necessity of structure and content from all types of nodes. 2) PC-GCN uses no extra information (the information of other posts in the same topic), the performance is still better than the baselines (Table 2 and 4), showing the effectiveness of our methods. 3) The models deleting comment information, i.e., TP-GCN and TP(RC)-GCN, experience a dramatic drop in performance, which shows the comment information is of the most importance. 4) The effect of structural information varies in the different situations. Without the contents, the comment structure can individually work (TP(RC)-GCN $>$ TP-GCN), while for topics, the structure has to collaborate with the contents ((RT)PC-GCN $<$ PC-GCN on the Weibo dataset).



\noindent\textbf{Ablation Study of DTPC-GCN}

We focus on the roles of the $U$ (topic-unrelated) branch and $R$ (topic-related) branch:

\textbf{U branch only}: Only $U$ branch is trained to capture topic-unrelated features.

\textbf{R branch only}: Only $R$ branch is trained to capture topic-related features.

Table \ref{ablation study:inter} shows that both of the two branches can identify controversial posts well, but their performances are worse than the fusion model. Specifically, the $U$ branch performs slightly better than $R$, indicating the topic-unrelated features are more suitable for inter-topic detection. We infer that the two branches can learn good but different representation under the guide of the auxiliary task.



\begin{figure}[!ht]
	\centering
	\includegraphics[width=0.48\textwidth]{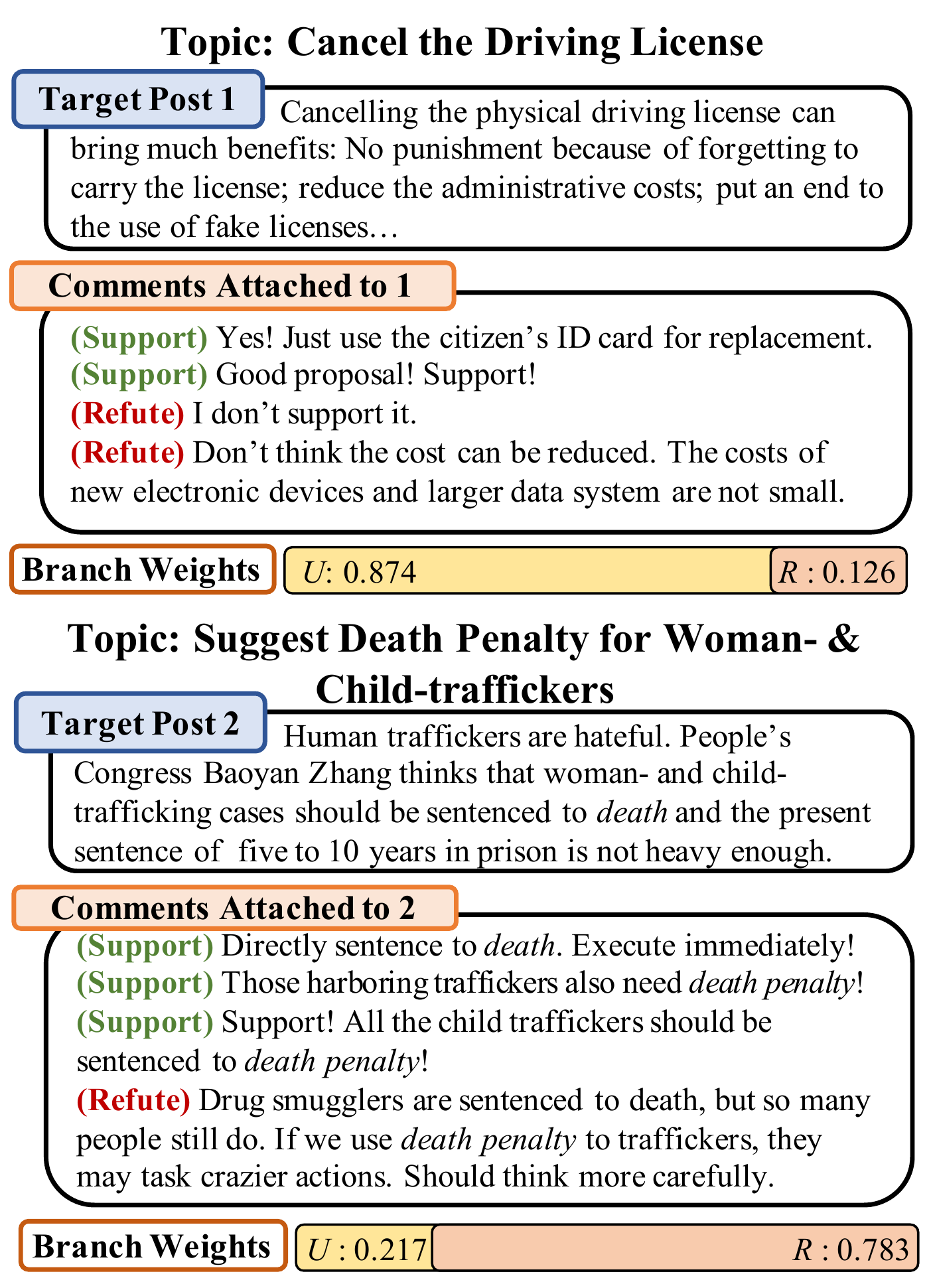}
	\caption{Examples of controversial posts that rely more on one of the two branches. The attention weights of the two posts are on the horizontal bars (left: the $U$ branch, right: the $R$ branch). Post $1$ rely more on $U$ ($0.874>0.126$) while Post $2$ more on $R$ ($0.217<0.783$).}
	\label{fig:casestudy}
\end{figure}

\subsection{Case Study}

We conduct a case study to further answer \textbf{EQ3} from the perspective of samples. We compare the attention weight of the $U$ and $R$ branch in DTPC-GCN and exhibit some examples where the final decisions lean on one of the two branches.

Figure \ref{fig:casestudy} shows two examples in the testing set of the Weibo dataset. The DTPC-GCN rely more on the topic-unrelated features from Branch $U$ when classifying Post $1$ ($0.874>0.126$), while more on the topic-related features from Branch $R$ when classifying Post $2$ ($0.217<0.783$). The topic of Post $1$, \textit{Cancel the Driving License}, is weakly relevant to topics in training set, and the comments mostly use topic-unspecific words such as simple \textit{support} and \textit{good proposal}. Thus, the topic-unrelated features are more beneficial for judging. In contrast, Post $2$ discusses the death penalty for women and children traffickers, relevant to one of the topics in the training set, \textit{Improve Sentencing Standards for Sexually Assault on Children}. Further, both of the two topics are full of comments on \textit{death penalty}. Exploiting more of the topic-related features is reasonable for the final decision.

\subsection{Error Analysis}

By conducting the error analysis on 186 misclassified samples in the Weibo dataset, we find three main types of samples that lead to the misclassification: 1) 22.6\% of the wrong samples are with too much noise in the comments, including unrelated and neutral comments. 2) 16.1\% are with a very deep tree structure. This kind of structure is helpful for controversy detection \cite{hessel2019something}, but the ability of GCN to obtain information from this kind of structure is limited.
3) 10.2\% are with obscure and complex statements.
These wrong cases indicate that better handling the noisy data, learning more deep structural features, and mining the semantic more deeply have the potential to improve the performance.

\section{Conclusion}
In this paper, we propose a novel method TPC-GCN to integrate the information from the graph structure and content of topics, posts, and comments for post-level controversy detection on social media. Unlike the existing works, we exploit the information from related posts in the same topic and the reply structure for more effective detection. To improve the performance of our model for inter-topic detection, we propose an extension of TPC-GCN named DTPC-GCN, to disentangle the topic-related and topic-unrelated features and then dynamically fuse them. Extensive experiments conducted on two datasets demonstrate that our proposed models outperform the compared methods and prove that our models can integrate both semantic and structural information with significant genaralizablity.

\section*{Acknowledgments}
The authors thank Peng Qi, Mingyan Lu, Guang Yang, and Jiachen Wang for helpful discussion.
This work is supported by the National Nature Science Foundation of China (U1703261). 



\bibliography{acl2020}
\bibliographystyle{acl_natbib}
\clearpage
\appendix


\section{Topics in the Weibo dataset}
\label{append:topics}
\begin{center}
\small
\setlength{\tabcolsep}{1pt}
\begin{tabular}{rp{0.95\textwidth}}
\hline \textbf{\#} & \textbf{Topics} \\ 
\hline
1 & Wechat businessman Ting Zhang and his wife paid taxes of 2.1 billion. (\begin{CJK*}{UTF8}{gbsn}张庭夫妇微商纳税21亿 \end{CJK*}) \\
2 & Singer Zhiqian Xue climbed a telegraph pole. (\begin{CJK*}{UTF8}{gbsn}薛之谦爬电线杆\end{CJK*}) \\
3 & Young Artist Yuan Wang was spotted to smoke. (\begin{CJK*}{UTF8}{gbsn}王源抽烟\end{CJK*}) \\
4 & Actor Yunlei Zhang believe women must do home cleaning well. (\begin{CJK*}{UTF8}{gbsn}张云雷 女人连家务活都不干好\end{CJK*}) \\
5 & Jiuxiang Sun sparred with the audience. (\begin{CJK*}{UTF8}{gbsn}孙九香怼观众\end{CJK*}) \\
6 & Host Xin Wu sold the gift that Actor Hanliang Zhong gave. (\begin{CJK*}{UTF8}{gbsn}吴昕将钟汉良送的礼物卖了\end{CJK*})\\
7 & Director Huatao Teng said he wrongly invited Actor Han Lu. (\begin{CJK*}{UTF8}{gbsn}滕华涛称用错了鹿晗\end{CJK*}) \\
8 & Actor Changjiang Pan responded for his not knowing who Xukun Cai was. (\begin{CJK*}{UTF8}{gbsn}潘长江回应不认识蔡徐坤\end{CJK*})\\
9 & Constume drama and idol drama will be off air from August. (\begin{CJK*}{UTF8}{gbsn}8月起停播娱乐性古装剧偶像剧\end{CJK*})\\
10 & A woman who was questioned to occupy the seats showed six train tickets. (\begin{CJK*}{UTF8}{gbsn}女子被质疑霸座掏出6张车票\end{CJK*}) \\
11 & An Internet user was detained for creating doggerels that slandered the Yichun City's image. (\begin{CJK*}{UTF8}{gbsn}打油诗 拘留\end{CJK*})\\
12 & Scanning QR codes can let you know the cleaning times of hotel sheets. (\begin{CJK*}{UTF8}{gbsn}酒店床单洗过几次扫码即知\end{CJK*})\\
13 & 31 names of places that do not conform the regulations in Xiamen are required to change. (\begin{CJK*}{UTF8}{gbsn}厦门31个不规范地名被要求整改\end{CJK*})\\
14 & Traditional Chinese medicine injection. (\begin{CJK*}{UTF8}{gbsn}中药注射液\end{CJK*})\\
15 & Jilin University provides wake-up services for foreign students. (\begin{CJK*}{UTF8}{gbsn}吉林大学为留学生提供叫醒服务\end{CJK*})\\
16 & A Gaokao-taking student who was rejected by Peking University for three times in the same year responded. (\begin{CJK*}{UTF8}{gbsn}考生回应被北大三次退档\end{CJK*})\\
17 & Xiaohongshu App was removed by top Android app stores. (\begin{CJK*}{UTF8}{gbsn}小红书疑被各大安卓应用商店下架\end{CJK*})\\
18 & FView questioned the authenticity of the Moon photo captured by the Huawei phone. (\begin{CJK*}{UTF8}{gbsn}爱否质疑华为拍的月亮造假\end{CJK*}) \\
19 & A new advertisement of Burger King is suspected of racial discrimination. (\begin{CJK*}{UTF8}{gbsn}汉堡王新广告被指种族歧视\end{CJK*})\\
20 & Zara responded for being suspected of uglifying a Chinese model. (zara\begin{CJK*}{UTF8}{gbsn}回应丑化中国模特\end{CJK*})\\
21 & A microblogger implied that Xiaomi’s Mimoji copied Apple’s Memoji. (\begin{CJK*}{UTF8}{gbsn}小米回应萌拍抄袭苹果事件\end{CJK*})\\
22 & Baidu CEO Robin Li was splashed water. (\begin{CJK*}{UTF8}{gbsn}李彦宏被泼水\end{CJK*})\\
23 & Huawei announced HarmonyOS.  (\begin{CJK*}{UTF8}{gbsn}华为鸿蒙系统发布\end{CJK*})\\
24 & Xiaomi adjusted its organizational structure. (\begin{CJK*}{UTF8}{gbsn}小米组织架构调整\end{CJK*})\\
25 & Resume the mandatory before-marriage examination. (\begin{CJK*}{UTF8}{gbsn}建议恢复强制性婚检\end{CJK*})\\
26 & Add another legal day-off every other week. (\begin{CJK*}{UTF8}{gbsn}建议每周双休改成隔周三休\end{CJK*})\\
27 & Lower the legal marriageable age to 20 for male and 18 for female. (\begin{CJK*}{UTF8}{gbsn}建议法定最低婚龄修订男20女18\end{CJK*})\\
28 & Cancel the driving license. (\begin{CJK*}{UTF8}{gbsn}建议取消机动车驾驶证\end{CJK*})\\
29 & Lower the minimum age of criminal responsibility for juveniles to 12. (\begin{CJK*}{UTF8}{gbsn}建议未成年人刑责年龄降至12岁\end{CJK*})\\
30 & The salary of teachers should not be lower than civil servants. (\begin{CJK*}{UTF8}{gbsn}教师待遇不应低于公务员\end{CJK*})\\
31 & Regulate the phenomenon that let parents check homework. (\begin{CJK*}{UTF8}{gbsn}建议严禁批作业转移给家长\end{CJK*})\\
32 & Suggest printing horror pictures on cigarette boxes. (\begin{CJK*}{UTF8}{gbsn}建议烟盒印恐怖图片\end{CJK*})\\
33 & Improve Sentencing Standards for Sexually Assault on Children (\begin{CJK*}{UTF8}{gbsn}完善性侵儿童犯罪量刑标准\end{CJK*})\\
34 & Suggest a minor long leave every month. (\begin{CJK*}{UTF8}{gbsn}建议实行每月一次小长假\end{CJK*})\\
35 & Women with a second child should have more supporting policies.  (\begin{CJK*}{UTF8}{gbsn}建议给予生二胎女性更多配套措施\end{CJK*})\\
36 & Suggest promoting education of death for all citizens. (\begin{CJK*}{UTF8}{gbsn}建议全民开展死亡教育\end{CJK*})\\
37 & Both of the wife and husband should have maternity leave. (\begin{CJK*}{UTF8}{gbsn}建议夫妻一起休产假\end{CJK*})\\
38 & Suggest extending women's maternity leave by one month. (\begin{CJK*}{UTF8}{gbsn}建议女性产假延长一个月\end{CJK*})\\
39 & Need heavier punishment to the violence to doctors. (\begin{CJK*}{UTF8}{gbsn}建议对暴力伤医从严判决\end{CJK*})\\
40 & Suggest at least 10 years in prison for child-traffickers. (\begin{CJK*}{UTF8}{gbsn}建议拐卖儿童最低刑期10年\end{CJK*})\\
41 & Suggest different prices for seat tickets and stand-by tickets. (\begin{CJK*}{UTF8}{gbsn}建议改进高铁站票座票同价\end{CJK*})\\
42 & Severely punish the juveniles for violating the law on purpose. (\begin{CJK*}{UTF8}{gbsn}建议严管未成年人知法犯法\end{CJK*})\\
43 & Suggest death penalty for woman- and child-traffickers. (\begin{CJK*}{UTF8}{gbsn}建议拐卖妇女儿童罪最高调至死刑\end{CJK*})\\
44 & The Double First-Class University list should be allowed to change.  (\begin{CJK*}{UTF8}{gbsn}建议双一流大学名单流动\end{CJK*})\\
45 & Forbid the no-dining-room catering companies to deliver take-out food. (\begin{CJK*}{UTF8}{gbsn}建议严禁无实体店外卖\end{CJK*}) \\
46 & Include the lunar New Year's Eve in the legal holidays. (\begin{CJK*}{UTF8}{gbsn}建议年三十纳入法定假期\end{CJK*})\\
47 & Give special care to menstrual female employees. 
(\begin{CJK*}{UTF8}{gbsn}建议给经期女职工特殊保护\end{CJK*})\\
48 & Forbid the juveniles' being live video streamers on the Internet. (\begin{CJK*}{UTF8}{gbsn}建议禁止未成年人担任网络主播\end{CJK*})\\
49 & Suggest parents going to schools for learning to be a qualified parents. (\begin{CJK*}{UTF8}{gbsn}建议上家长学校学当家长\end{CJK*})\\
\hline
\end{tabular}

\begin{minipage}[b]{0.95\textwidth}
\makeatletter\def\@captype{table}\makeatother 
\caption{\label{table:49topics} 49 topics in the Weibo dataset. We modify some words and polish the sentences to improve the understandability when translating them into English.}
\end{minipage}

\end{center}

\end{document}